\documentclass[11pt,a4paper]{article}
\usepackage[hyperref]{conll-2019}
\usepackage{times}
\usepackage{latexsym}

\usepackage{times}  
\usepackage{helvet}  
\usepackage{courier}  
\usepackage{url}  
\usepackage{graphicx}  
\usepackage{tikz}
\usepackage{bm}
\usepackage{relsize}
\usepackage{pgfplots}

\usepackage{graphicx}
\usepackage{epstopdf}
\usepackage{tabularx}
\usepackage{xcolor}
\usepackage{multirow}
\usepackage{bbm}
\usepackage[flushleft]{threeparttable}
\usepackage{amsmath}
\usepackage{amsfonts}
\usepackage{booktabs}
\usepackage{subcaption}

\usetikzlibrary{arrows,shapes, decorations.pathmorphing,backgrounds,positioning}

\aclfinalcopy 

\title{Variational Semi-supervised Aspect-term Sentiment Analysis via Transformer}

\makeatletter
\newcommand{\thankssymb}[1]{%
  \textsuperscript{\@fnsymbol{#1}}%
}
\makeatother

\author{Xingyi Cheng\thankssymb{1}, Weidi Xu~\thanks{\thankssymb{1}: equal contribution}, Taifeng Wang, Weipeng Huang, Kunlong Chen \and Wei Chu \\
	Ant Financial Services Group \\
	{\tt \{fanyin.cxy,weidi.xwd,taifeng.wang,weipeng.hwp,kunlong.ckl,weichu.cw\}@antfin.com}}

\date{}

\begin{document}
\maketitle
\begin{abstract}
	 Aspect-term sentiment analysis (ATSA) is a long-standing challenge in natural language processing. It requires fine-grained semantical reasoning about a target entity appeared in the text.
	 As manual annotation over the aspects is laborious and time-consuming, the amount of labeled data is limited for supervised learning.
	 This paper proposes a semi-supervised method for the ATSA problem by using the Variational Autoencoder based on Transformer. The model learns the latent distribution via variational inference. By disentangling the latent representation into the aspect-specific sentiment and the lexical context, our method induces the underlying sentiment prediction for the unlabeled data, which then benefits the ATSA classifier.
	 Our method is classifier-agnostic, i.e., the classifier is an independent module and various supervised models can be integrated.
	 Experimental results are obtained on the SemEval 2014 task 4 and show that our method is effective with different five specific classifiers and outperforms these models by a significant margin.
 \end{abstract}

\section{Introduction}

Aspect based sentiment analysis (ABSA) has two sub-tasks, namely aspect-term sentiment analysis (ATSA) and aspect-category sentiment analysis (ACSA). ACSA is to infer the sentiment polarity with regard to the predefined categories, e.g., the aspect $food, price, ambience$. On the other hand, ATSA aims at classifying the sentiment polarity of a given aspect word or phrase in the text. 
For example, given a review about a restaurant ``\emph{the $[pizza]_{aspect}$ is the best if you like thin crusted pizza, however, the $[service]_{aspect}$ is awful.}'', the sentiment implications with regard to ``\emph{pizza}'' and ``\emph{service}'' are contrary.
For the aspect ``\emph{pizza}'', the sentiment polarity is ``\emph{positive}'' while ``\emph{negative}'' for the aspect ``\emph{service}''.
In contrast to document-level sentiment analysis, ATSA requires more fine-grained reasoning about the textual context.
The task is worthy of investigation as it can obtain the attitude with regard to a specific entity which we are interested in. The task is widely applicated in analyzing the comments, such as opinion generation.
Recently, many attempts ~\cite{DBLP:conf/emnlp/TangQL16,DBLP:conf/acl/PanW18,DBLP:conf/naacl/LiuCB18,xue2018aspect,li2018hierarchical} focus on supervised learning and pay much attention to the interaction between the aspect and the context. However, the amount of labeled data is quite limited as the annotation about the aspects is laborious.
Currently available data sets, e.g. SemEval, only has around 2K unique sentences and 3K sentence-aspect pairs, which is insufficient to fully exploit the power of the deep models. 
Fortunately, a large amount of unlabeled data is available for free and can be accessed easily from the websites.
It will be of great significance if numerous unlabeled samples can be utilized to further facilitate the supervised ATSA classifier.
Therefore, the semi-supervised ATSA is a promising research topic.

In ATSA, achieving the sentiment of the aspect-term is semantically complicated and it is non-trivial for a model to capture sentimental similarity of the aspects, which causes the difficulties for semi-supervised learning. In this paper, we proposed a classifier-agnostic framework which named Aspect-term Semi-supervised Variational Autoencoder~\cite{kingma2013auto} based on Transformer (ASVAET). The variational autoencoder offers the flexibility to customize the model structure. In other words, the proposed framework is compatible with other supervised neural networks to boost their performance. Our proposed model learns the latent representation of the input data and disentangles the representations into two independent parts, i.e., the aspect-term sentiment and the representation of the lexical context. By regarding the aspect sentiment polarity of the unlabeled data as the discrete latent variable, the model implicitly induces the sentiment polarity via the variational inference.
Specifically, the representation of the lexical context is extracted by the encoder and the aspect-term sentiment polarity is inferred from the specific ATSA classifier.
The decoder takes these two representations as inputs and reconstructs the original sentence by two unidirectional language models. 
In contrast to the conventional auto-regressive models, the latent representations have their specific meanings and are obtained from the encoder and the classifier to the input examples. Therefore, it is also possible to condition the sentence generation on the sentiment and lexical information w.r.t. a certain target entity.
In addition, by separating the representation of the input sentence, the classifier becomes an independent module in our framework, which endows the method with the ability to integrate different classifiers.
The method is presented in detail in Sec.~\ref{sec:model}.

Experimental results are obtained on the two classical datasets from SemEval 2014 task 4~\cite{DBLP:conf/semeval/PontikiGPPAM14}.
Five recent available models are implemented as the classifier in ASVAET.
Our method is able to utilize the unlabeled data and consistently improve the performance against the supervised models.
Compared with other semi-supervised methods, i.e., in-domain word embedding pre-training and self-training, the proposed method also demonstrates better performance. We also evaluate the effectiveness of labeled data and sharing embeddings, and show that the structure can provide the separation between lexical context and sentiment polarity in the latent space.


\section{Related Work}
Sentiment analysis is a traditional research hotspot in the NLP field~\cite{wang2012baselines}.
Rather than obtaining the sentimental inclination of the entire text, ATSA instead aims to extract the sentimental expression w.r.t. a target entity.
With the release of online completions, abundant methods were proposed to explore the limits of current models.
Tang et al.~\cite{DBLP:conf/coling/TangQFL16} proposed to make use of bidirectional Long Short-Term Memory (LSTM)~\cite{DBLP:journals/neco/HochreiterS97} to encode the sentence from the left and right to the aspect-term.
This model primarily verifies the effectiveness of deep models for ABSA
Tang et al.~\cite{DBLP:conf/emnlp/TangQL16} then put forward a neural reasoning model in analogy to the memory network to perform the reasoning in many steps.
There are also many other works dedicating to solve this task~\cite{DBLP:conf/acl/PanW18,DBLP:conf/naacl/LiuCB18,DBLP:conf/eacl/ZhangL17}.

Another related topic is semi-supervised learning for the text classification. Recently, Data augmentation methods~\cite{xie2019unsupervised,berthelot2019mixmatch} achieve a greate success on low-resource datasets. Moreover,
A simple but efficient method is to use pre-trained modules, e.g., initializing the word embedding or bottom layers with pre-training. Word embedding technique has been wildly used in NLP models, e.g., Glove~\cite{DBLP:conf/emnlp/PenningtonSM14} and ELMo~\cite{DBLP:conf/naacl/PetersNIGCLZ18}. Recently, Bidirectional Encoder Representations from Transformer (BERT)~\cite{devlin2018bert} replaces the embedding layer to context-dependent layer with the pre-trained bidirectional language model to capture the contextual representation. BERT is complementary to the encoder of the proposed method. To keep our framework neat, these pre-training investigations are not conducted in this paper.

VAE-based semi-supervised methods, on the other hand, are able to cooperate with various kinds of classifiers.
VAE has been applied in many semi-supervised NLP tasks, ranging from text classification~\cite{xu2017variational}, relation extraction~\cite{DBLP:journals/tacl/MarcheggianiT16} to sequence tagging~\cite{chen2018variational}. 
Different from text classification where sentiment polarity is related to an entire sentence, ATSA just interested in related information of a given aspect-term. To circumvent this problem, a novel structure is proposed.

\section{Method Description}
\label{sec:model}
In this section, the problem definition is provided and then the model framework is presented in detail.

The ATSA task aims to classify a data sample with input sentence $\mathbf{x}=\{x_1, ..., x_n\}$ and corresponding aspect~\footnote{If an input sentence has $n$ aspect-terms, then $n$ data samples are generated.} $\mathbf{a}=\{a_1, ..., a_m\}$, where $\textbf{a}$ is a subsequence of $\textbf{x}$, into a sentiment polarity $y$, where $y \in \{P, O, N\}$.
$P, O, N$ denotes ``positive", ``neutral", ``negative".
For the semi-supervised ATSA, we consider the following scenario.
Given a dataset consisting of labeled samples $\mathbf{S}_l$ and unlabeled samples $\textbf{S}_u$, where the  $\mathbf{S}_l=\{(\mathbf{x}_l^{(i)},\mathbf{a}_l^{(i)}, y_l^{(i)})\}^{N_l}_{i=1}$ and $\mathbf{S}_u=\{(\mathbf{x}_u^{(i)},\mathbf{a}_u^{(i)}\}^{N_l}_{i=1}$, the goal is to utilize $\mathbf{S}_u$ to improve the classification performance over the supervised model using $\mathbf{S}_l$ only.

\begin{figure*}
	\centering
	\includegraphics[width=6.3in]{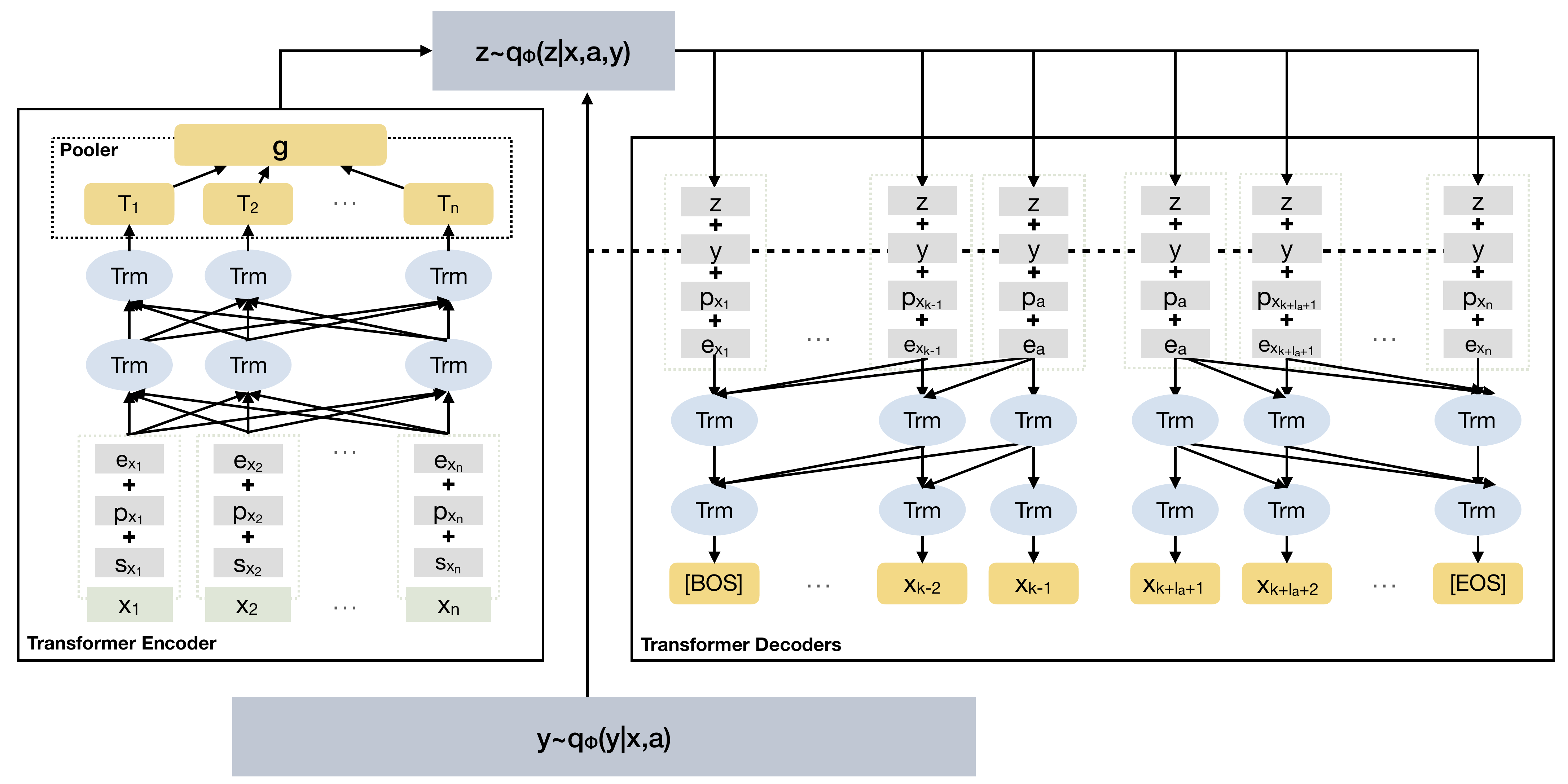}
	\caption{This is the sketch of our model with bidirectional encoder and decoder. Assuming the aspect-term starts at the $k$-th position in $\mathbf{x}$.
		{\bf Bottom}: When using unlabeled data, the distribution of $y\sim q_\phi(y|\mathbf{x}, \mathbf{a})$ is provided by the classifier.
		{\bf Left}: The sequence is encoded by a Transformer block, which receives the summation of three embeddings, i.e., segment (used to distinguish aspect words) $\mathbf{s}_{x_n}$, position $\mathbf{p}_{x_n}$ and word $\mathbf{e}_{x_n}$. The encoding and the label $y$ are used to parameterize the posterior $q_\phi(z|\textbf{x},\mathbf{a},y)$.
		{\bf Right}: A sample $\mathbf{z}$ from the posterior $q_\phi(\mathbf{z}|\mathbf{x}, \mathbf{a}, y)$ and label $y$ are passed to the generative network which estimates the probability $p_\theta(\mathbf{x}|y, \mathbf{a}, \mathbf{z})$ by two unidirectional Transformer decoders. The number of aspect tokens is $l_a$.
	}
	\label{fig:framework}
\end{figure*}

The architecture is depicted in Fig.~\ref{fig:framework}.
The method consists of three main components, i.e., the classifier, the encoder, and the decoder.	
The classifier can be any differentiable supervised ATSA model, which takes $\mathbf{x}$ and $\mathbf{a}$ as input, and outputs the prediction about $y$.
The encoder transform the data into a latent space that is independent of the label $y$.
And the decoder combines the outputs from the classifier and the encoder to reconstruct the input sentence.
For the labeled data, the classifier and the autoencoder are trained with the given label $y$.
For the unlabeled data, the $y$ is regarded as the latent discrete variable and it is induced by maximizing the generative probability.
As the classifier can be implemented by various models, the description of the classifier will be omitted.
We present a autoencoder structure based on Transformer~\cite{DBLP:conf/nips/VaswaniSPUJGKP17}. In the following, the objective functions are clarified, followed by the model description.

\subsection{Variational Inference}
\label{sec:vi}
\begin{figure}
	\centering
	\includegraphics[width=2.0in]{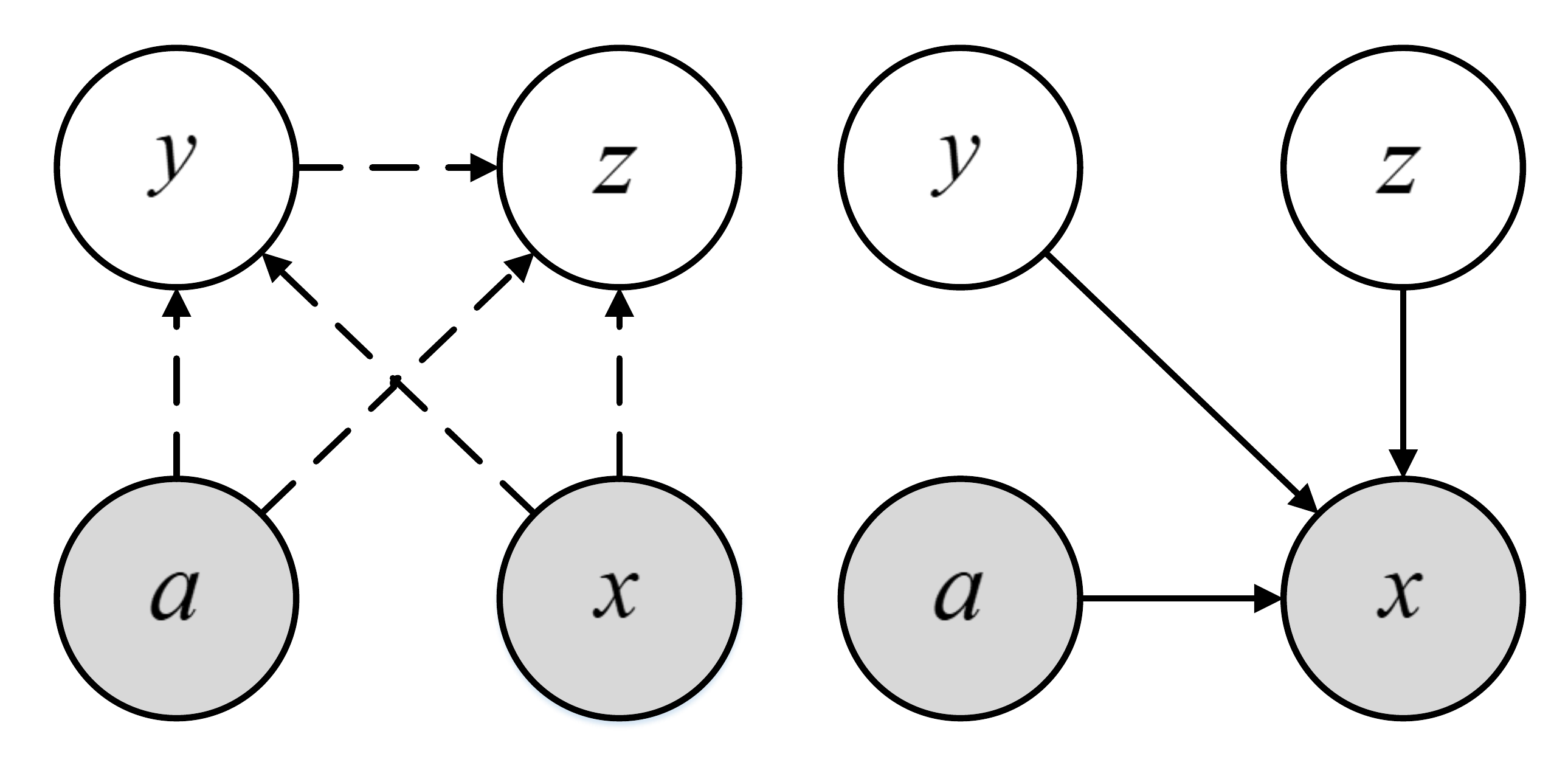}
	\caption{Illustration of ASVAET as a directed graph. \textbf{Left}: Dashed lines are used to denote variational approximation $q_\phi(y|\mathbf{x},\mathbf{a})q_\phi(\mathbf{z}|\mathbf{x},\mathbf{a}, y)$. \textbf{Right}: Solid lines are used to denote generative model $p_\theta(\mathbf{x}|y,\mathbf{a},\mathbf{z})$.}\label{fig:vi}
\end{figure}

Using generative models is a common approach for semi-supervised learning, which tries to extract the information from the unlabeled data by modeling the data distribution.
In VAE, the data distribution is modeled by optimizing the evidence lower bound (ELBO) of data log-likelihood, which leads to two objectives for labeled data and unlabeled data respectively.
For the labeled data, VAE maximizes the ELBO of $p(\mathbf{x},y|\mathbf{a})$. For the unlabeled data, it optimizes the ELBO of $p(\mathbf{x}|\mathbf{a})$, where the $y$ is latent and integrated.
Specifically, the dependency between variables is illustrated in Fig.~\ref{fig:vi}.
The ELBO of $\log p(\mathbf{x},y|\mathbf{a})$ can be given as follows:
\begin{eqnarray}
\begin{split}
\log p_{\theta}(\mathbf{x}, y|\mathbf{a}) & \ge \mathbb{E}_{q_{\phi}(\mathbf{z}|\mathbf{x},\mathbf{a},y)}[\log p_{\theta}(\mathbf{x}|y, \mathbf{a}, \mathbf{z})] \\ 
& - D_{KL}(q_{\phi}(\mathbf{z}|\mathbf{x},\mathbf{a},y) || p_\theta(\mathbf{z})) \\
& + \log p_\theta(y) \\
& = \mathcal{L}(\mathbf{x}, \mathbf{a}, y) \,,
\end{split}
\label{equ:lab}
\end{eqnarray}
where $\mathbf{z}$ is the latent variable which represents lexical information over the sentence and $D_{KL}$ is the Kullback–Leibler divergence.

In terms of the unlabeled data, the ELBO of $\log p(\mathbf{x}|\mathbf{a})$ can be extended from Eq.~\ref{equ:lab}.
\begin{eqnarray}
\begin{split}
\log p_{\theta}(\mathbf{x}|\mathbf{a}) & \ge \sum_{y}q_{\phi}(y|\mathbf{x}, \mathbf{a})(\mathcal{L}(\mathbf{x}, \mathbf{a}, y)) \\ 
& + \mathcal{H}(q_{\phi}(y|\mathbf{x}, \mathbf{a})) \\
& = \mathcal{U}(\mathbf{x}, \mathbf{a})\,,
\end{split}
\label{equ:unl}
\end{eqnarray}
where $\mathcal{H}$ is the entropy function and $q_{\phi}(y|\mathbf{x}, \mathbf{a})$ is the classification function.

And $q_{\phi}(y|\mathbf{x}, \mathbf{a})$ can also be trained in the supervised manner using the labeled data.
Combining the above objectives, the overall objective for the entire data set is:

\begin{eqnarray}
\begin{split}
J & = \sum_{(\mathbf{x}, \mathbf{a}, y) \in S_{l}}-\mathcal{L}(\mathbf{x}, \mathbf{a}, y) + \sum_{\mathbf{x} \in S_{u}}-\mathcal{U}(\mathbf{x}, \mathbf{a}) \\
& + \gamma \sum_{(\mathbf{x}, \mathbf{a}, y) \in S_{l}}-\log q_{\phi}(y|\mathbf{x}, \mathbf{a}) \,,
\end{split}
\label{equ:com}
\end{eqnarray}
where  $\gamma$ is a hyper-parameter which controls the weight of the additional classification loss. 

To implement this objective, three components are required to model the $q_{\phi}(y|\mathbf{x}, \mathbf{a})$, $q_{\phi}(\mathbf{z}|\mathbf{x},\mathbf{a},y)$ and $p_{\theta}(\mathbf{x}|y, \mathbf{a}, \mathbf{z})$ respectively.

\subsection{Classifier}
Various currently available models can be used as the classifier. 
For the unlabeled data, the classifier is used to predict the distribution of label $y$ for the decoder, i.e., $y \sim q_{\phi}(y|\mathbf{x}, \mathbf{a})$.
The distribution $q_{\phi}(y|\mathbf{x}, \mathbf{a})$ will be tuned during maximizing the objective in Eq.~\ref{equ:unl}.
In this work, five classifiers are implemented in ASVAET and they are also used as the supervised baselines for the comparison.

\subsection{Transformer Encoder}
The encoder plays the role of $q_{\phi}(\mathbf{z}|\mathbf{x},\mathbf{a},y)$.
This module attempts to extract the lexical feature that is independent of the label $y$ when given data sample $(\mathbf{x}, \mathbf{a})$.
In this way, the $\mathbf{z}$ and $\mathbf{y}$ jointly form the representative vector for the input data.

In our implementation, we use a bidirectional encoder to construct sentences embeddings. It is referred as the Transformer encoder that is actually a sub-graph of the Transformer architechture~\cite{DBLP:conf/nips/VaswaniSPUJGKP17}, the architecture is shown in the left part of the Fig.~\ref{fig:vi}. The encoder employs residual connections
around each of the multi-head attention sub-layers, followed by layer normalization. To capture the aspect-term, we treat the aspect-term and its context differently by segment embeddings. To further emphasize the position of the conditional aspect, the position tag is also included for each token.
The position tag indicates the distance from the token to the aspect.
And then the position tag is transformed into a vector as defined in ~\cite{DBLP:conf/nips/VaswaniSPUJGKP17}, which is added with the word embedding and segment embedding as the input of the Transformer encoder.
Let $\mathbf{g}$ denote the output of the Transformer encoder after pooling which simply averaging the hidden states of the aspect-terms (the number of tokens is equal or greater than one) of the last layer, $\mathbf{y}$ is the indicator vector of the polarity. Then the distribution of $\mathbf{z}$ can be given as:
\begin{align}
	\mathbf{z} & \sim \mathcal{N}(\mu(\mathbf{x},y), diag(\sigma^2(\mathbf{x},y)))\,, \nonumber\\
	\mu(\mathbf{x},y) &= \tanh(\mathbf{W}_\mu [\mathbf{g}:\mathbf{y}] + \mathbf{b}_\mu) \,,\nonumber\\
	\sigma(\mathbf{x},y) &= \tanh(\mathbf{W}_\sigma [\mathbf{g}:\mathbf{y}] + \mathbf{b}_\sigma) \,.\nonumber
\end{align}

The sequences are divided into two parts by using segment embedding, the encoder can be aware of the position and the content of the aspect-term $\mathbf{a}$ by multi-head attention operation in the Transformer encoder.
The information from two sides are aggregated into the aspect-term $\mathbf{a}$, and therefore the resulting $\mathbf{z}$ can gather the information related to the aspect.


\subsection{Transformer Decoders}
The decoder is also a sub-graph of Transformer architechture~\cite{DBLP:conf/nips/VaswaniSPUJGKP17} which focus on reconstructing original text. The main difference from the Transformer encoder is that the Transformer decoder is unidirectional by modifying the self-attention sub-layer to prevent positions from paying attention to subsequent positions. The textual sequence is well-known to be semantically complex and it is non-trivial for a Transformer decoder to capture the high-level semantics.
Here we investigate two questions.
How to implement $p_\theta(\mathbf{x}| y,\mathbf{a}, \mathbf{z})$ without losing the information of $\mathbf{a}$ and how to capture the semantic polarity by a sequential model. For the first question, denoting that $\mathbf{x}$ is composed of three parts $(\mathbf{x}_l, \mathbf{a}, \mathbf{x}_r)$, we use two Transformer decoders to model the left and right content.
For the second question, we let each token is generated conditioned on the summation of the variables $\mathbf{z}$ and embedding $\mathbf{y}$.


One way to achieve $p_\theta(\mathbf{x}|y, \mathbf{a}, \mathbf{z})$ is to separate the sequence into two parts, reversing the process in the two unidirectional decoder.
For each decoder, the input state is represented by the summation of the four input i.e., the polarity indicator vector $\mathbf{y}$ from the classifier or the labeled dataset, the context vector $\textbf{z}$ from the encoder, the input token embedding $\mathbf{e}_{x_t}$ and the position embedding $\mathbf{p}_{x_t}$:
\begin{align}
	\overleftarrow{\mathbf{h}}^{trm}_t = \overleftarrow{f_{trm}}(\mathbf{e}_{[x_t:a]}, \mathbf{p}_{x_t}, \mathbf{y}, \mathbf{z}), \quad x_t \in [\mathbf{x}_l : \mathbf{a}] \nonumber\\
	p({x}_{t-1}|\cdot) = \text{softmax}(\textbf{W}_p \overleftarrow{\textbf{h}}^{trm}_t + b_p) \,, \nonumber\\
	\log p_\theta(\mathbf{x}_l|\mathbf{a}, {y}, \mathbf{z}) = \sum_{x_t} \log p(x_t|\cdot), \quad x_t \in \mathbf{x}_l \,, \nonumber\\
	\overrightarrow{\textbf{h}}^{trm}_t = \overrightarrow{f_{trm}}(\mathbf{e}_{[a:x_t]}, \mathbf{p}_{x_t}, \mathbf{y}, \mathbf{z}), \quad x_t \in [\mathbf{a} : \mathbf{x}_r] \nonumber\\
	p({x}_{t+1}|\cdot) = \text{softmax}(\textbf{W}_p \overrightarrow{\textbf{h}}^{trm}_t + b_p) \,, \nonumber\\
	\log p_\theta(\mathbf{x}_r|\mathbf{a}, {y}, \mathbf{z}) = \sum_{x_t} \log p(x_t|\cdot), \quad x_t \in \mathbf{\mathbf{x}_r}  \,.\nonumber
\end{align}
It is equivalent to generate two sequences using two decoders.
When decoding left part (or right part), the aspect will first get processed by the decoder and hence the decoder is aware of the aspect-terms.
The position tag is also used in the decoder.

\section{Experiments}
\subsection{Datasets and Preparation}
The models are evaluated on two benchmarks: Restaurant (\texttt{REST}) and Laptop (\texttt{LAPTOP}) datasets from the SemEval ATSA challenge~\cite{DBLP:conf/semeval/PontikiGPPAM14}.
The \texttt{REST} dataset contains the reviews in the restaurant domain, while the \texttt{LAPTOP} dataset contains the reviews of Laptop products.
The statistics of these two datasets are listed in Table ~\ref{tab:sta}.
When processing these two datasets, we follow the same procedures as in another work~\cite{DBLP:conf/acl/LamLSB18}.
The dataset has a few samples that are labeled as ``conflict'' and these samples are removed.
All tokens in the samples are lowercased without other preprocess, e.g., removing the stop words, symbols or digits.

In terms of the unlabeled data, we obtained samples in the same domain for the \texttt{REST} and \texttt{LAPTOP} datasets.
For the \texttt{REST}, the unlabeled samples are obtained from a sentiment analysis competition in Kaggle~\footnote{https://inclass.kaggle.com/c/restaurant-reviews}.
The competition consists of 82K training samples and 34K test samples.
For the \texttt{LAPTOP}, the unlabeled samples are obtained from the ``Six Categories of Amazon Product Reviews"~\footnote{http://times.cs.uiuc.edu/~wang296/Data/}, which has 412K samples.
The reviews about the laptops are used among six product categories.

The NLTK sentence tokenizer is utilized to extract the sentences from the raw comments.
And each sentence is regarded as a sample in our model for both \texttt{REST} and \texttt{LAPTOP}.
To obtain the aspects in the unlabeled samples, an open-sourced aspect extractor~\footnote{https://github.com/guillaumegenthial/sequence\_tagging} is pre-trained using labeled data.
The resulting test F1 score is 88.42 for the \texttt{REST} and 80.12 for the \texttt{LAPTOP}.
Then the unlabeled data is processed by the pre-trained aspect extractor to obtain the aspects.
The sentences that have no aspect are removed.
And the sentences are filtered with maximal sentence length 80.
The  statistic of the resulting sentences is given in Table.~\ref{tab:ssta}.

\begin{table}
	\small
	\begin{tabular}{c c c c c}
		\toprule
		&   & \# Positive & \# Negative & \# Neutral \\
		\hline
		\multirow{ 2}{*}{ \texttt{REST}} & Train & 2159 & 800 & 632 \\
		& Test & 730 & 195 & 196   \\
		\hline
		\multirow{ 2}{*}{ \texttt{LAPTOP}} & Train & 980 & 858 & 454 \\
		& Test & 340 & 128 & 171   \\
		\toprule
	\end{tabular}
	\caption{The statistics of the datasets.}
	\label{tab:sta}
\end{table}

\begin{table}
	\small
	\centering
	\begin{tabular}{c c c c}
		\toprule
		&   & Avg. Length &  Std. Length  \\
		\hline
		\multirow{ 2}{*}{ \texttt{REST}} & Labeled & 20.06 & 10.38  \\
		& Unlabeled & 22.70 & 12.38   \\
		\hline
		\multirow{ 2}{*}{ \texttt{LAPTOP}} & Labeled & 21.95 & 11.80  \\
		& Unlabeled & 29.89 & 17.33   \\
		\toprule
	\end{tabular}
	\caption{The statistics of the reviews.}
	\label{tab:ssta}
\end{table}

\begin{table*}
	\small
	\centering
	\resizebox{0.8\linewidth}{!}{
		\begin{tabular}{l l l l l l }
			\toprule
			\multirow{2}{*}{Classifier} & \multirow{2}{*}{Models} & \multicolumn{2}{c}{\texttt{REST}} &  \multicolumn{2}{c}{\texttt{LAPTOP}} \\
			\cline{3-6}
			& & Accuracy & Macro-F1 & Accuracy & Macro-F1 \\
			\hline
			- & CNN-ASP & 77.82 $\natural$ & - & 72.46 $\natural$ & - \\
			- & AE-LSTM & 76.60 $\natural$ & - & 68.90 $\natural$ & - \\
			- & ATAE-LSTM & 77.20 $\natural$ & - & 68.70 $\natural$ & - \\
			- & GCAE & 77.28 (0.32) $\natural$ & - & 69.14 (0.32) $\natural$ & - \\
			\hline
			\multirow{3}{*}{TC-LSTM}
			& TC-LSTM & 77.97 (0.16) & 67.55 (0.32) & 68.42 (0.56)& 62.42 (1.10)\\
			& TC-LSTM (EMB) & 77.18 (0.38) & 65.97 (0.44) & 67.51 (0.72) & 60.31 (1.28)\\
			& TC-LSTM (ST) & 78.19 (0.36) & 67.65 (0.43) & 68.47 (0.47) & 62.54 (0.74) \\
			& TC-LSTM (ASVAET) & \textbf{78.34} (0.18) & \textbf{68.41} (0.92) & \textbf{70.04} (0.53) & \textbf{64.23} (0.71)\\
			\hline
			\multirow{3}{*}{MemNet}
			& MemNet & 78.68 (0.23) & 68.18 (0.58) & 70.28 (0.32) & 64.38 (0.86)\\
			& MemNet (EMB) & 79.47 (0.38) & 69.06 (0.21) & 72.17 (0.44) & 65.06 (0.73)\\
			& MemNet (ST) & 78.83 (0.20) & 68.92 (0.20) & 69.52 (0.36) & 64.39 (0.67) \\
			& MemNet (ASVAET) & \textbf{80.58} (0.23) & \textbf{70.06} (0.53) & \textbf{73.21} (0.55)& \textbf{65.88} (0.45) \\
			\hline
			\multirow{3}{*}{IAN}
			& IAN & 79.20 (0.19) & 68.71 (0.59) & 69.48 (0.52) & 62.90 (0.99) \\
			& IAN (EMB) & 79.46 (0.38) & 69.45 (0.38) & 70.89 (0.48) & 65.27 (0.34) \\
			& IAN (ST) & 79.45 (0.11) & 69.36 (0.71) & 73.25 (0.81) & 68.25 (0.76) \\
			& IAN (ASVAET) & \textbf{80.23} (0.17) & \textbf{70.32} (1.00) & \textbf{74.02} (0.42) & \textbf{69.39} (0.75) \\ 
			\hline
			\multirow{3}{*}{BILSTM-ATT-G}
			& BILSTM-ATT-G & 79.74 (0.22) & 69.16 (0.53) & 74.26 (0.35) & 69.54 (0.53) \\
			& BILSTM-ATT-G (EMB) & 80.27 (0.44) & 70.33 (0.51) & 73.61 (0.30) & 68.25 (0.63) \\
			& BILSTM-ATT-G (ST) & 80.54 (0.23) & 71.88 (0.19) & 74.70 (0.41)  & 70.31 (0.60)\\
			& BILSTM-ATT-G (ASVAET) & \textbf{81.11} (0.34) & \textbf{72.19} (0.27)  & \textbf{75.44} (0.32)  &  \textbf{70.52} (0.33)\\
			\hline
			\multirow{3}{*}{TNet-AS}
			& TNet-AS & 80.56 (0.23) & 71.17 (0.43) & 76.75 (0.35) & 71.88 (0.35) \\
			& TNet-AS (EMB) & 80.96 (0.49) & 69.99 (0.87) & 76.45 (0.40) & 71.52 (0.73) \\
			& TNet-AS (ST) & 80.76 (0.23) & 71.32 (0.56) & 76.88 (0.41)  & 71.74 (0.63)\\
			& TNet-AS (ASVAET) &\textbf{81.77} (0.20) & \textbf{72.57} (0.32) & \textbf{77.57} (0.31) & \textbf{72.31} (0.69) \\
			\hline
		\end{tabular}
	}
	\caption{Experimental results (\%). For each classifier, we performed five experiments, i.e., the supervised classifier, the supervised classifier with pre-trained embedding using unlabeled data and our model with the classifier. The results are obtained after 5 runs, and we report the mean and the standard deviation of the test accuracy, and the Macro-averaged F1 score. Better results are in bold. $\natural$ denotes that the results are extracted from the original paper.}
	\label{tab:results}
\end{table*}

\subsection{Model Configuration \& Classifiers}
In the experiments, the model is fixed with a set of universal hyper-parameters.
The number of units in the encoder and the decoder is 100 and the latent variable is of size 50 and the number of layers of both Transformer blocks is 2, the number of self-attention heads is 8.
The KL weight $klw$ should be carefully tuned to prevent the model from trapping in a local optimum, where $\mathbf{z}$ carries no useful information.
In this work, the KL weight is set to be 1e-4.
In term of word embedding, the pre-trained GloVe~\cite{DBLP:conf/emnlp/PenningtonSM14} is used as the input of the encoder and the decoder~\footnote{http://nlp.stanford.edu/data/glove.8B.300d.zip} and the out-of-vocabulary words are excluded.
And it is fixed during the training.
The $\gamma$ is set to be 10 across the experiments.


We implemented and verified four kinds of mainstream ATSA classifiers integrated into our model, i.e., TC-LSTM~\cite{DBLP:conf/coling/TangQFL16}, MemNet~\cite{DBLP:conf/emnlp/TangQL16}, BILSTM-ATT-G~\cite{DBLP:conf/eacl/ZhangL17}, IAN~\cite{DBLP:conf/ijcai/MaLZW17} and TNet~\cite{li2018transformation}.
\begin{itemize}
	\item \textbf{TC-LSTM}: Two LSTMs are used to model the left and right context of the target separately, then the concatenation of two representations is used to predict the label.
	\item \textbf{MemNet}: It uses the attention mechanism over the word embedding over multiple rounds to aggregate the information in the sentence, the vector of the final round is used for the prediction.
	\item \textbf{IAN}: IAN adopts two LSTMs to derive the representations of the context and the target phrase interactively and the concatenation is fed to the softmax layer.
	\item \textbf{BILSTM-ATT-G}: It models left and right contexts using two attention-based LSTMs and makes use of a special gate layer to combine these two representations. The resulting vector is used for the prediction.
		\item \textbf{TNet-AS}: Without using an attention module, TNet adopts a convolutional layer to get salient features from the transformed word representations originated from a bi-directional LSTM layer. Among current supervised models, TNet is currently one of the in-domain state-of-the-art methods and the TNet-AS is one of the two variants of TNet. 
	
\end{itemize}
The configuration of hyper-parameters and the training settings are the same as in the original papers.
Various classifiers are tested here to demonstrate the robustness of our method and show that the performance can be consistently improved for different classifiers.

\subsection{Main Results}

Table~\ref{tab:results} shows the experimental results on the \texttt{REST} and \texttt{LAPTOP} datasets.
Two evaluation metrics are used here, i.e., classification accuracy and Macro-averaged F1 score.
The latter is more sensitive when the dataset is class-imbalance.
In this table, the semi-supervised results are obtained with 10K unlabeled data.
We didn't observe further improvement with more unlabeled data.
The mean and the standard deviation are reported over 5 runs. 
For each classifier \emph{clf}, we conducted the following experiments:
\begin{itemize}
	\item \emph{clf}: The classifier is trained using labeled data only.
	\item \emph{clf} (EMB): We use CBOW~\cite{DBLP:journals/corr/abs-1301-3781} to train the word embedding vectors using both labeled and unlabeled data. And the resulting vectors, instead of pre-trained GloVe vectors, are used to initialize the embedding matrix of the classifier. This is the embedding-level semi-supervised learning as the embedding layer is trained using in-domain data.\
	\item \emph{clf} (ST): The self-training (ST) method is a typical semi-supervised learning method. We performed the self-training method over each classifier. At each epoch, we select the 1K samples with the best confidence and give them pseudo labels using the prediction. Then the classifier is re-trained with the new labeled data. The procedure loops until all the unlabeled samples are labeled.
	\item \emph{clf} (ASVAET): The proposed method that uses \emph{clf} as the classifier. Note again that the classifier is an independent module in the proposed model, and the same configuration is used as in the supervised learning. 
\end{itemize}

Besides, we also include the results of several supervised models in the first block, i.e., CNN-ASP~\cite{DBLP:conf/acl/LamLSB18}, AE-LSTM, ATAE-LSTM~\cite{DBLP:conf/emnlp/WangHZZ16}, GCAE~\cite{DBLP:conf/acl/LiX18}, from the original paper.

From the Table~\ref{tab:results}, the ASVAET is able to improve supervised performance consistently for all classifiers. For the MemNet, the test accuracy can be improved by about 2\% by the TSSVAE, and so as the Macro-averaged F1. The TNet-AS outperforms the other three models.

Compared with the other two semi-supervised methods, the ASVAET also shows better results.
The ASVAET outperforms the compared semi-supervised methods evidently. The adoption of in-domain pre-trained word vectors is beneficial for the performance compared with the Glove vectors. 


\subsection{Ablation Studies}
\subsubsection{Effect of Labeled Data}
\begin{figure}
	\centering
	\includegraphics[height=1.8in]{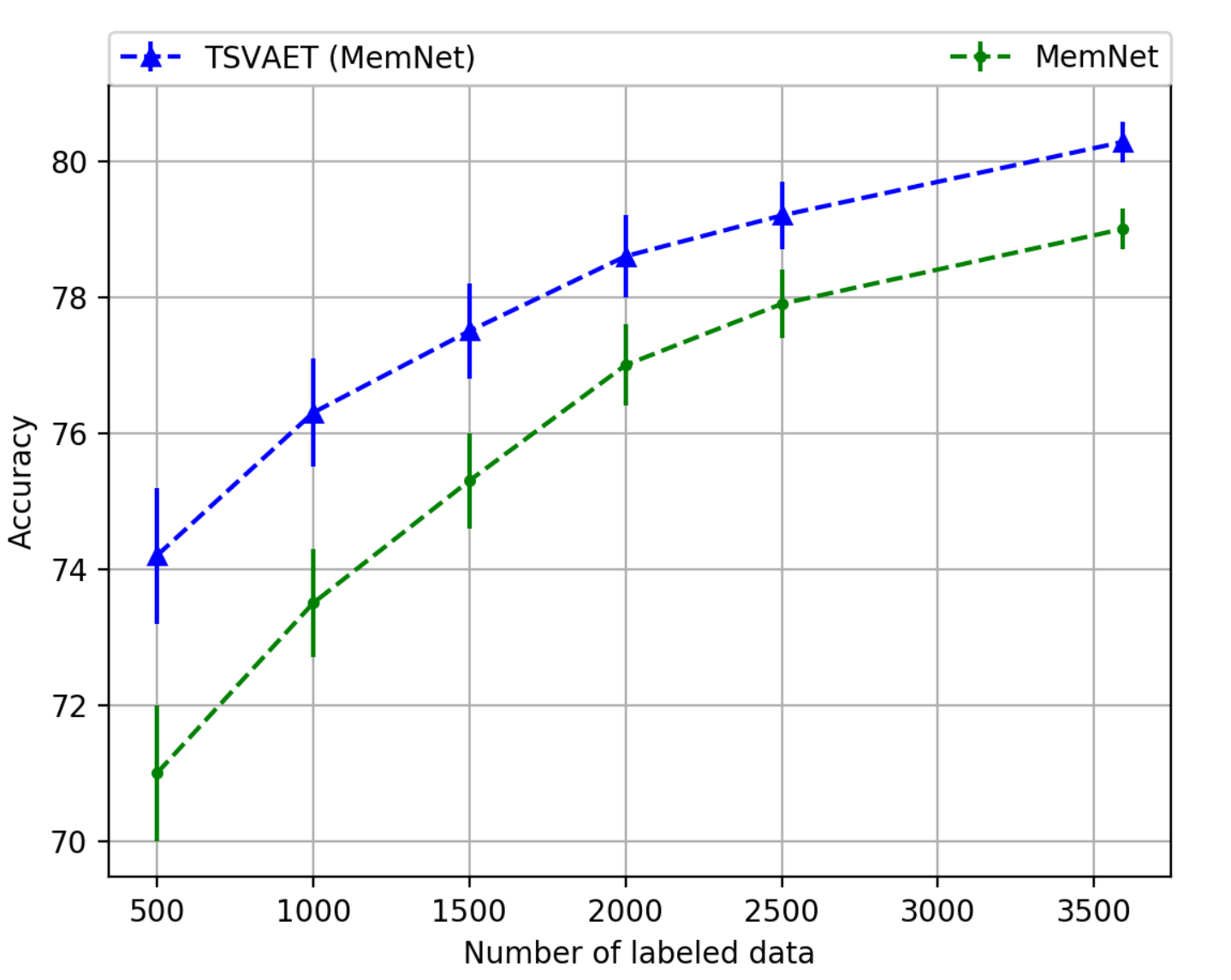}
	\caption{The test accuracy w.r.t. the number of labeled samples on the \texttt{REST} dataset with MemNet classifier.}\label{fig:accwrtlabel}
	\centering
	\includegraphics[height=1.8in]{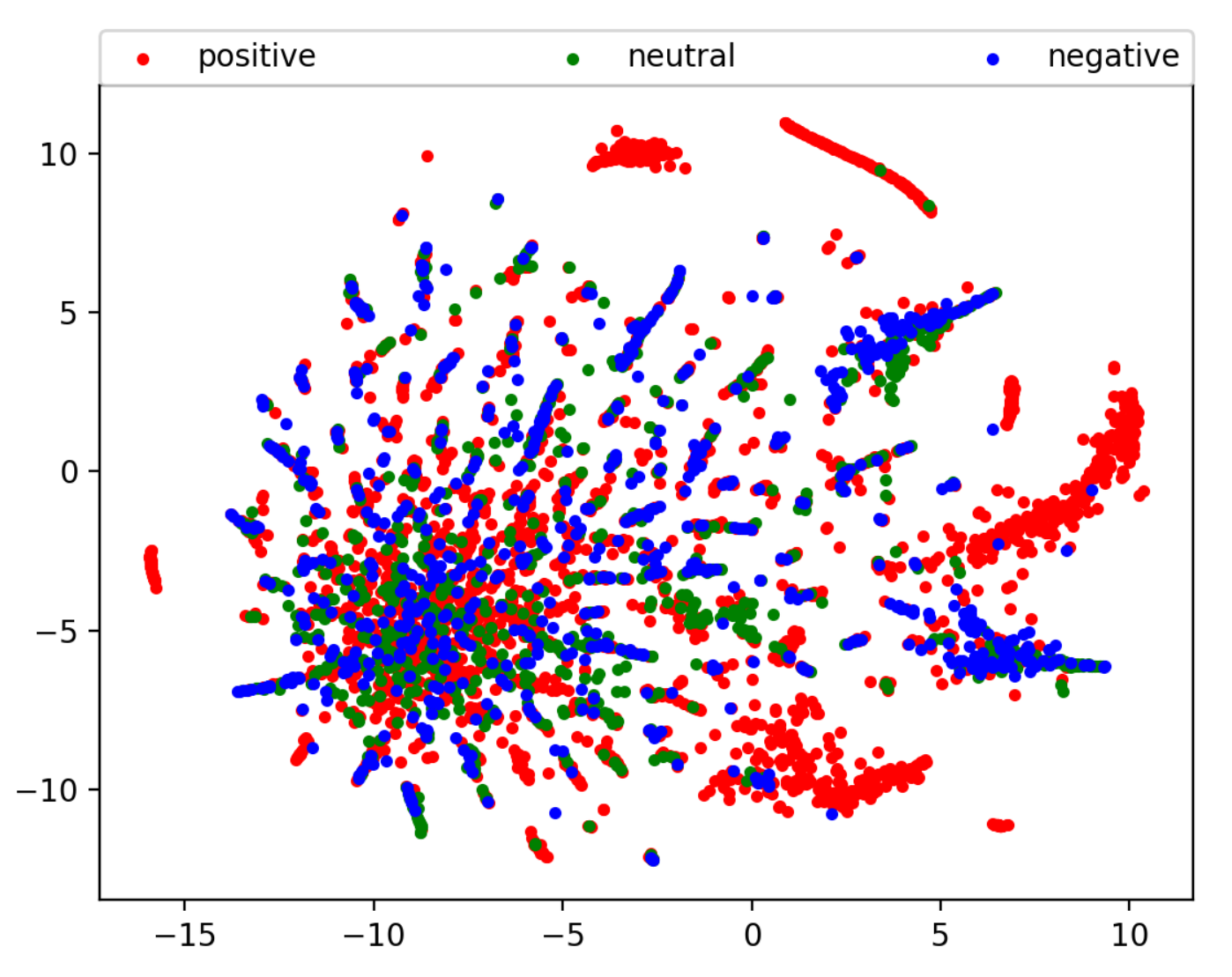}
	\caption{The distribution of the \texttt{REST} dataset in latent space $\mathbf{z}$ using t-SNE.}\label{fig:tsne}
\end{figure}

Here we investigated whether the ASVAET works with less labeled data.
Without loss of generality,  the MemNet is used as the basic classifier.
We sampled different amount of labeled data to verify the improvement by using ASVAET.
The test accuracy curve w.r.t. the amount of labeled data used is shown in Fig.~\ref{fig:accwrtlabel}.
With fewer labeled samples, the test accuracy decreases, however, the improvement becomes more evident.
When using 500 labeled samples, the improvement is about 3.2\%.
With full 3591 labeled samples, 1.5\% gain can be obtained.
This illustrates that our method can improve the accuracy with limited data.

\begin{table}
	\small
	\centering
	\begin{tabular}{c c c}
		\toprule
		Accuracy  & w/o sharing & w/ sharing \\
		\hline
		TC-LSTM (ASVAET) & \textbf{78.34} & 77.65 \\
		MemNet (ASVAET) & \textbf{80.58} & 78.82 \\
		IAN (ASVAET) & \textbf{80.23} & 79.22\\
		BILSTM-ATT-G (ASVAET) & \textbf{81.11} & 78.36 \\
		TNet-AS (ASVAET) & \textbf{81.77} & 79.53 \\
		\toprule
	\end{tabular}
	\caption{Comparison between with or without sharing embedding on the \texttt{REST} dataset.}\label{tab:emb}
\end{table}

\subsubsection{Effect of Sharing Embeddings}
In previous works, the word embedding is shared among
all the components. In other words, the word embedding is also tuned in learning to reconstruct the data. It is questionable whether the improvement is obtained by using VAE or multi-task learning (text generation and classification). In the aforementioned experiments, the embedding layer is not shared between the classifier and autoencoder. This implementation guarantees that the improvement does not come from learning to generate.
To verify if sharing embedding will benefit, we also conducted experiments with sharing embedding, as illustrated in Table.~\ref{tab:emb}.
The results indicate that the joint training for the embedding layer is negative for improving the performance in this task. The gradient from the autoencoder may collide with the gradients from the classifier and therefore,
interferes with the optimization direction.

\begin{table}
	\centering
	\begin{tabular}{ccc}
		\toprule
		\textbf{Positive}\\
		\hline
		... the best \emph{food} i 've ever had !!! ...  \\
		... the \emph{lox} is very tasty ... \\
		... the \emph{rice} is a great value ... \\
		\toprule
		\textbf{Negative}\\
		\hline
		... the worst \emph{food} i 've ever had !!! ... \\
		... the \emph{lox} is a bit of boring ... \\
		... the \emph{rice} is awful ... \\
		\toprule
		\textbf{Neutral}\\
		\hline
		... had the \emph{food} in the restaurant ... \\
		... lox with a glass of chilli sauce ... \\
		... the \emph{rice} with a couple of olives salad ... \\
		\hline
	\end{tabular}
	\caption{Nice sentences that are generated by controlling the sentiment polarity $y$ using the decoder. }\label{tab:gen}
\end{table}

\subsection{Analysis of the Latent Space}
Transformer encodes the data into two representations, i.e., $y$ and $\mathbf{z}$. These two latent variable represented sentiment polarity and lexical context individually from the input text.
We expect the $y$ and $\mathbf{z}$ are fully disentangled and represent different meanings. The scatters of latent variable $\mathbf{z}$ (cf. Fig.~\ref{fig:tsne}) helps us have a better understanding.
As shown in the figure, the distributions of three different polarities are very similar, which indicates that the lexical context reprensetation $\mathbf{z}$ is independent of the polarity $y$.

The generation ability of the decoder is also investigated.
Several sentences are generated and selected in the Table~\ref{tab:gen}.
By controlling the sentiment polarity $y$ with the same $\mathbf{z}$, the decoder can generate sentences with different sentiment in a similar format.
This indicates that the decoder is trained successfully to perceive the $y$ and model the relationship between the $y$ and $\mathbf{x}$.




\section{Conclusion}
A VAE-based framework has been proposed for the ATSA task.
In this work, the encoder and decoder are constructed from the Transformers.  Both analytical and experimental work has been carried out to show the effectiveness of the ASVAET.
The method is verified with various kinds of classifiers.
For all tested classifiers, the improvement is obtained when equipped with ASVAET, which demonstrates its universality.

In this paper, the aspect-term is assumed to be known and there is an error accumulation problem when using the pre-trained aspect extractor. According to this, in future work, it is also interesting to show if it is possible to learn the aspect and sentiment polarity jointly for the unlabeled data. It will be of great importance if detailed knowledge can be extracted from the unlabeled data, which will shed light on other related tasks.

\bibliography{conll-2019}
\bibliographystyle{acl_natbib}

\end{document}